\documentclass[runningheads]{llncs}
\usepackage{amssymb}
\usepackage{comment}
\usepackage{fancyvrb}
\usepackage[title]{appendix}
\usepackage[usenames,dvipsnames]{xcolor}
\usepackage{subfig}
\usepackage{array}
\usepackage{lscape}
\usepackage{hyperref}

\usepackage{lipsum}   
\usepackage{setspace} 

\usepackage{etoolbox}
\AtBeginEnvironment{quote}{\singlespacing\small}

\usepackage[T1]{fontenc}
\usepackage{graphicx}
\begin{document}
\title{\textit{Valet}: A Standardized Testbed of Traditional Imperfect-Information Card Games}
\titlerunning{\textit{Valet}: A Standardized Testbed of Traditional Card Games}
%
\author{Mark Goadrich\inst{1}\orcidID{0000-0002-9137-7836} \and
Achille Morenville\inst{2}\orcidID{0009-0003-4014-8181} \and
Éric Piette\inst{2}\orcidID{0000-0001-8355-636X}}
\authorrunning{M. Goadrich, A. Morenville, and É. Piette}
%
\institute{Hendrix College, Conway AR 72032, USA\\
\email{goadrich@hendrix.edu} \and
ICTEAM, UCLouvain, Louvain-la-Neuve, Belgium\\
\email{\{achille.morenville,eric.piette\}@uclouvain.be}}
\maketitle              
\begin{abstract}
AI algorithms for imperfect-information games are typically compared using performance metrics on individual games, making it difficult to assess robustness across game choices. Card games are a natural domain for imperfect information due to hidden hands and stochastic draws. To facilitate comparative research on imperfect-information game-playing algorithms and game systems, we introduce \textit{Valet}, a diverse and comprehensive testbed of 21 traditional imperfect-information card games. These games span multiple genres, cultures, player counts, deck structures, mechanics, winning conditions, and methods of hiding and revealing information. To standardize implementations across systems, we encode the rules of each game in \textit{RECYCLE}, a card game description language. We empirically characterize each game's branching factor and duration using random simulations, reporting baseline score distributions for a Monte Carlo Tree Search player against random opponents to demonstrate the suitability of \textit{Valet} as a benchmarking suite.

\end{abstract}

\section{Introduction}
Tabletop card games have been played worldwide for centuries and remain a rich domain for studying decision-making under uncertainty. Unlike perfect-information games, they naturally incorporate hidden hands, stochastic draws, and information revelation through play, requiring players to reason about beliefs, deception, and risk. These properties make imperfect-information card games a compelling benchmark for Game AI, yet evaluations are often performed on only a few games, leaving the effect of game choice on reported performance largely unexplored. Card games are a canonical class of imperfect-information games \cite{blair1993games}, in which players cannot observe the full game state because key information is hidden in private hands and revealed progressively as play unfolds. This partial observability makes traditional game-playing techniques such as alpha-beta pruning \cite{knuth1975analysis} difficult to apply directly, motivating specialized approaches that reason under uncertainty, including Monte Carlo Tree Search \cite{swiechowski2023monte} and Counterfactual Regret Minimization \cite{zinkevich2007regret}.

In empirical evaluations of imperfect-information game-playing algorithms, researchers have primarily focused on a small set of individual games, often drawn from three recurring genres of card games: trick-taking games (such as \textsc{Hearts} \cite{goadrich2021quantifying}, \textsc{Spades} \cite{baier2018emulating}, \textsc{Doppelkopf} \cite{sievers2015doppelkopf}, \textsc{Skat} \cite{buro2009improving}, \textsc{Jass} \cite{niklaus2019survey}, and
\textsc{Bridge} \cite{ginsberg2001gib}), climbing games (such as
\textsc{Dou Di Zhu} \cite{whitehouse2011determinization} and
\textsc{Big 2} \cite{charlesworth2018application}),
and betting games (most commonly variants of \textsc{Poker} \cite{brown2019deep}). In many cases, the choice of game is justified by regional popularity or simply taken for granted. As a result, the extent to which reported performance depends on the selected game remains unclear, motivating the need for a systematic understanding of how game mechanics influence algorithm behaviour.

Several general game-playing (GGP) and game AI frameworks support research on imperfect-information games through collections of implemented domains, including card games, but none provides a shared set implemented consistently across systems. RLCard \cite{zha2019rlcard} is a reinforcement-learning toolkit, offering ten games across six genres, while OpenSpiel \cite{LanctotEtAl2019OpenSpiel} provides a broader suite for game-theoretic and reinforcement-learning research, including seventeen card games alongside many non-card domains. CardStock \cite{bell2016automated} provides over 70 traditional and modern card games encoded in \textit{RECYCLE} \cite{bell2016automated}, and supports agents based on determinization within Monte Carlo methods. TAG \cite{gaina2020tag} similarly targets comparative research across modern board and card games, including fifteen card games, and Belief-Stochastic Game Framework (BSG) \cite{morenville2024belief,morenville2025cog} centralizes belief modelling in the game engine and includes five games, four of which are card games. Despite this progress, cross-system comparisons remain difficult: no single game is implemented in all frameworks, and overlapping games often differ in rules or conventions. This motivates the need for a shared and diverse testbed for imperfect-information card games.

To support comparative research on imperfect-information game-playing algorithms and systems, we introduce \textit{Valet}, a diverse testbed of fixed rule-sets for twenty-one traditional card games, developed in the context of the GameTable COST Action \cite{Piette_2024_GameTable} to facilitate collaboration and shared evaluation practices. \textit{Valet} enables systematic comparisons across different GGP and game AI frameworks by providing a common suite of games with varying information structures and stochasticity. It also supports fairer and more reproducible agent evaluation by reducing reliance on small, selective benchmark sets, thereby limiting the influence of game selection and allowing performance differences to be attributed more confidently to algorithmic design choices.

In this paper, we first detail the curation of the games in \textit{Valet}, describing their diversity in genre and cultural heritage, and how we stabilize their rule sets within a card game description language. Next, we present an empirical analysis demonstrating how the games vary in information flow branching factor, game length, and scoring distributions. We conclude by outlining avenues for future work that use \textit{Valet} to advance research on imperfect-information games.


\begin{landscape}
\bgroup
\setlength\extrarowheight{2pt} 
\setlength\tabcolsep{1pt}
\begin{table}[tp]
\caption{\textit{Valet} includes twenty-one games that represent a variety of genres, the number of players, deck compositions, mechanics, and winning conditions, along with different methods of hiding and revealing information.\\}\label{tab1}
\begin{tabular}{l|l|l|c|c|l|l|l|c|c|c}
\multicolumn{1}{c|}{\textbf{Game}} & \multicolumn{1}{c|}{\textbf{Genre}} & \multicolumn{1}{c|}{\textbf{Origin}} & \textbf{Date} & \textbf{Players} & \multicolumn{1}{c|}{\textbf{Deck}} & \multicolumn{1}{c|}{\textbf{Scoring}} & \textbf{Information} & \textbf{Tricks} & \textbf{Sets/Runs} & \textbf{Teams} \\ \hline
Agram                              & Last Trick                          & Niger                                & 2000            & 3                & Unique                             & One Winner                            & P, D            & \checkmark       &                    &                \\
BlackJack                          & Banking                             & France                               & 1930          & 1                & French                             & High Score                            &                  &                 &                    &                \\
Crazy Eights                       & Shedding                            & USA                                  & 1940          & 3                & French                             & Low Score                             & P               &                 &                    &                \\
Cribbage                           & Adding                              & England                              & 1600          & 2                & French                             & High Score                            & P, T, S, D         &                 & \checkmark          &                \\
Cuckoo                             & Exchange                            & France                              & 1490          & 6                & French                             & One Loser                             & P, S, D            &                 &                    &                \\
Euchre                             & Euchre                              & USA                                  & 1820          & 4                & Unique                             & High Score                            & P, T, S, D      & \checkmark       &                    & \checkmark      \\
Go Fish                            & Quartet                             & USA                                  & 1850          & 4                & French                             & High Score                            & P, T, D         &                 & \checkmark          &                \\
Golf-6                             & Draw \& Discard                     & USA                                  & ??            & 4                & French                             & Low Score                             & P               &                 &                    &                \\
Goofspiel                          & Collect                             & USA                                  & 1930          & 2                & French                             & High Score                            & P, B            &                 &                    &                \\
Hearts                             & Avoidance                           & USA                                  & 1880          & 4                & French                             & Low Score                             & P, D            & \checkmark       &                    &                \\
Klaverjassen                       & Jack-Nine                           & Netherlands                          & 1890          & 4                & Piquet                             & High Score                            & P, D            & \checkmark       & \checkmark          & \checkmark      \\
Leduc Hold'em                      & Poker                               & Canada                               & 2005          & 2                & Unique                             & High Score                            & P               &                 &                    &                \\
Pitch                              & High-Low-Jack                       & England                              & 1800          & 4                & French                             & High Score                            & P, D            & \checkmark       &                    & \checkmark      \\
President                          & Climbing                            & China                                & 1960          & 5                & French                             & High Score                            & P               &                 & \checkmark          &                \\
Rummy                              & Rummy                               & Mexico                                  & 1900          & 2                & French                             & Low Score                             & P, T            &                 & \checkmark          &                \\
Scarto                              & Tarot                               & Italy                                  & ??          & 3                & Tarot                             & High Score                             & P, D            &  \checkmark                &          &                \\
Schwimmen                          & Commerce                            & Austria                              & 1718            & 5                & Piquet                             & High Score                            & P, T, S         &                 & \checkmark          &                \\
Scopa                              & Fishing                             & Italy                                & 1700          & 2                & Italian                            & High Score                            & P               &                 &                    &                \\
Skitgubbe                          & Beating                             & Sweden                               & 1949            & 3                & French                             & One Loser                             & P, D            &                 &                    &                \\
Sueca                              & Ace-Ten                             & Portugal                             & 1800          & 4                & Spanish                            & High Score                            & P, D            & \checkmark       &                    & \checkmark      \\
Whist                              & Trick Taking                        & England                              & 1880          & 4                & French                             & High Score                            & P, D            & \checkmark       &                    & \checkmark     
\end{tabular}
\end{table}
\egroup
\end{landscape}

\section{Testbed Curation}

The \textit{Valet} testbed\footnote{\href{0}{https://mgoadric.github.io/valet/}}
is composed of twenty-one traditional card games, listed in Table \ref{tab1}. McLeod \cite{pagat} distinguishes between traditional (popular in specific regions with traditional card decks), invented (played with traditional decks but without a wide following), and commercial (which require purchasing a specialized deck) card games. We focus our work on traditional card games, as their mechanics are often foundations of modern commercial card games. 

We used the categorizations and classifications from McLeod’s \textit{Pagat Card Game Rules} website \cite{pagat} and Parlett’s \textit{The Penguin Book of Card Games} \cite{parlett2008penguin} to guide our selection of distinct genres, mechanics, and player counts. Our goal in constructing \textit{Valet} was to be comprehensive in our genre selection and to include, within each genre, a game that:
\begin{enumerate}
\item exhibits the defining qualities of the genre;
\item promotes diversity in cultural connections and play experiences;
\item reflects earlier examples in the historical development of the genre;
\item uses a comparatively simple rule set.
\end{enumerate}

As a consequence, we omit some widely studied classics, such as \textsc{Bridge}, \textsc{Skat}, and \textsc{Texas Hold'em}, and instead include related games such as \textsc{Whist}, \textsc{Klaverjassen}, and \textsc{Leduc Hold'em}. 

\subsection{Genre}

The games in \textit{Valet} span several major families of traditional card games. Trick-taking games are the most represented category, where players play one card each into a trick and the winner of the trick is determined by suit-following constraints and (in many cases) a trump suit. \textit{Valet} includes eight trick-taking games spanning distinct subgenres: \textsc{Agram}, \textsc{Whist}, \textsc{Euchre}, \textsc{Sueca}, \textsc{Hearts}, \textsc{Pitch}, \textsc{Klaverjassen}, and \textsc{Scarto}. These games vary substantially in scoring method (counting tricks, specific cards), objectives (high-score, low-score, and winner-takes-all), the use of partnerships, and how trump is selected.

A second group of games in \textit{Valet} centers on hand management and shedding mechanics, where players aim to reduce their hand size by playing valid cards or combinations. This category includes \textsc{Crazy Eights}, \textsc{President}, \textsc{Go Fish}, \textsc{Rummy}, and \textsc{Skitgubbe}, which range from simple matching rules to games requiring multi-step planning and information inference.

We also include four games based on hand comparison and exchange, where players attempt to end the game with a stronger hand than their opponents, often through limited swapping or betting-like decisions: \textsc{BlackJack}, \textsc{Leduc Hold'em}, \textsc{Cuckoo}, and \textsc{Schwimmen}. Finally, \textit{Valet} contains four games that do not fit cleanly into the above categories but contribute additional mechanics, including simultaneous play (\textsc{Goofspiel}), fishing and capture (\textsc{Scopa}), draw-and-discard with hidden ownership (\textsc{Golf-6}), and multi-stage scoring with distinct play phases (\textsc{Cribbage}).

\subsection{Deck Composition}

\textit{Valet} includes games that originated in
Austria, Canada, China, England, France, Italy, Mexico, The Netherlands, Niger, Portugal, Sweden, and the United States. This cultural diversity can be demonstrated by examining the variety of decks used by these games. The most common deck used in \textit{Valet} is the French-suited deck of 52 cards, which includes thirteen cards each in the suits Clubs, Diamonds, Hearts, and Spades, each with ranks from ace to ten, plus face cards of Jack, Queen, and King. The Tarot deck used in \textsc{Scarto} expands this deck to 78 cards, with four suits of fourteen cards, plus twenty-one separate cards in a Trump suit and one card called The Fool.
The Spanish and Italian-suited Decks, used by \textsc{Sueca} and \textsc{Scopa} respectively, consist of only 40 cards in four suits, each with ranks ace to seven and three face cards.
\textsc{Agram}, \textsc{Euchre}, \textsc{Klaverjassen}, and \textsc{Schwimmen} all use even smaller decks derived from the 
French-suited deck, while \textsc{Leduc Hold'em} employs the smallest deck with only six cards, with two each of rank King, Queen, and Jack.

\subsection{Variants and \textit{RECYCLE} Encoding}

Consistency is difficult in research involving card games, where rules are often modified and reinterpreted by local communities, leading to countless variants. Additionally, many games have been adapted to be played with a variable number of players. To stabilize the games in \textit{Valet}, the rules of each game, for a specific number of players, have been encoded in the \textit{RECYCLE} card game description language.\footnote{\href{1}{https://cardstock.readthedocs.io/en/latest/recycle/index.html}} We believe these encodings will provide a common reference for the alignment of implementations. 

Many card games include rules to repeatedly play until one player or team has reached a certain number of points. This has the effect of averaging out the randomness of the deal and can help better determine human skill. However, for the games in \textit{Valet}, we limit games when possible to a single round, thus making the decision space more approachable for AI players. All games except \textsc{Cribbage}, where we include two rounds of play, with each player taking a turn as dealer to balance out the dealer's advantage of scoring from discarded cards. 

We detail here the specific variants used in games in \textit{Valet}. We selected the popular American version of \textsc{Euchre}, which does not use the Joker, and we use the ``stick the dealer'' variant, forcing play each deal. In \textsc{Hearts} and \textsc{Scarto}, we do not include the ability to exchange or pass cards before playing the game. For \textsc{BlackJack}, the Ace can rank either high or low. We include limited betting options with insurance, but do not include an option to split hands. So the game ends in a reasonable amount of time, we chose the Block variant of \textsc{Rummy}, where the game concludes when the draw deck is exhausted. For a similar reason, our version of \textsc{Crazy Eights} requires players to always play a card if they have a match. We chose in \textsc{Goofspiel} to resolve ties by holding cards over to the next round.

\section{Evaluation}

To assess the relevance and diversity of the selected games, we 
investigate four key aspects: information flow, branching factor, game length, and score distribution. Both the branching factor and game length reveal the decision-making complexity of the game. Furthermore, distinct results for all four metrics across all games suggest diversity among them. 

All results are obtained from simulations conducted using the CardStock GGP system \cite{bell2016automated}. For each game, we run 200 simulations, 100 with all players making random choices, and 100 with a MCTS first player and all other players making random choices. For the MCTS player, we follow \cite{whitehouse2011determinization} such that agent creates 10 determinizations of the game state at each decision point, then uses a budget of 100 times the number of choices at each decision point to explore the search space within each determinization. 

\subsection{Information Flow}

\bgroup
\setlength\extrarowheight{1pt} 
\setlength\tabcolsep{3pt}

\begin{table}[t]
\caption{Card games hide and reveal information in different ways.}\label{tab2}
\begin{tabular}{l|l|l}
\textbf{Name} & \textbf{Type} & \textbf{Description} \\
\hline
Public	& Visibility & Cards are played face up, seen by all players. \\
Hidden	& Visibility & Cards are played face down, unseen by players. \\
Private	& Visibility &Cards are held by players to create private information. \\
Back	& Visibility &Cards are distinguished by type, indicated on the back. \\
Taken	& Movement &Public cards are moved to private or hidden locations. \\
Shared	& Movement &Private cards are moved to private or hidden locations. \\
Deduction	& Action &Knowledge of cards is implied by actions.  \\
\end{tabular}
\end{table}
\egroup

Table \ref{tab2} summarizes seven mechanisms by which card games hide and reveal information. Imperfect information arises naturally from card visibility: cards may be face up (public), face down (hidden), or held in players’ hands (private), and distinct back designs can indicate separate decks. In CardStock, visibility is attached to card locations, so all cards in a location share the same visibility; information is revealed or concealed as cards move between locations and ownership changes. Finally, some information is conveyed implicitly through player actions. For example, in trick-taking games, failing to follow suit reveals that a player holds no remaining cards of that suit.

\begin{figure}[t]
\includegraphics[width=\textwidth]{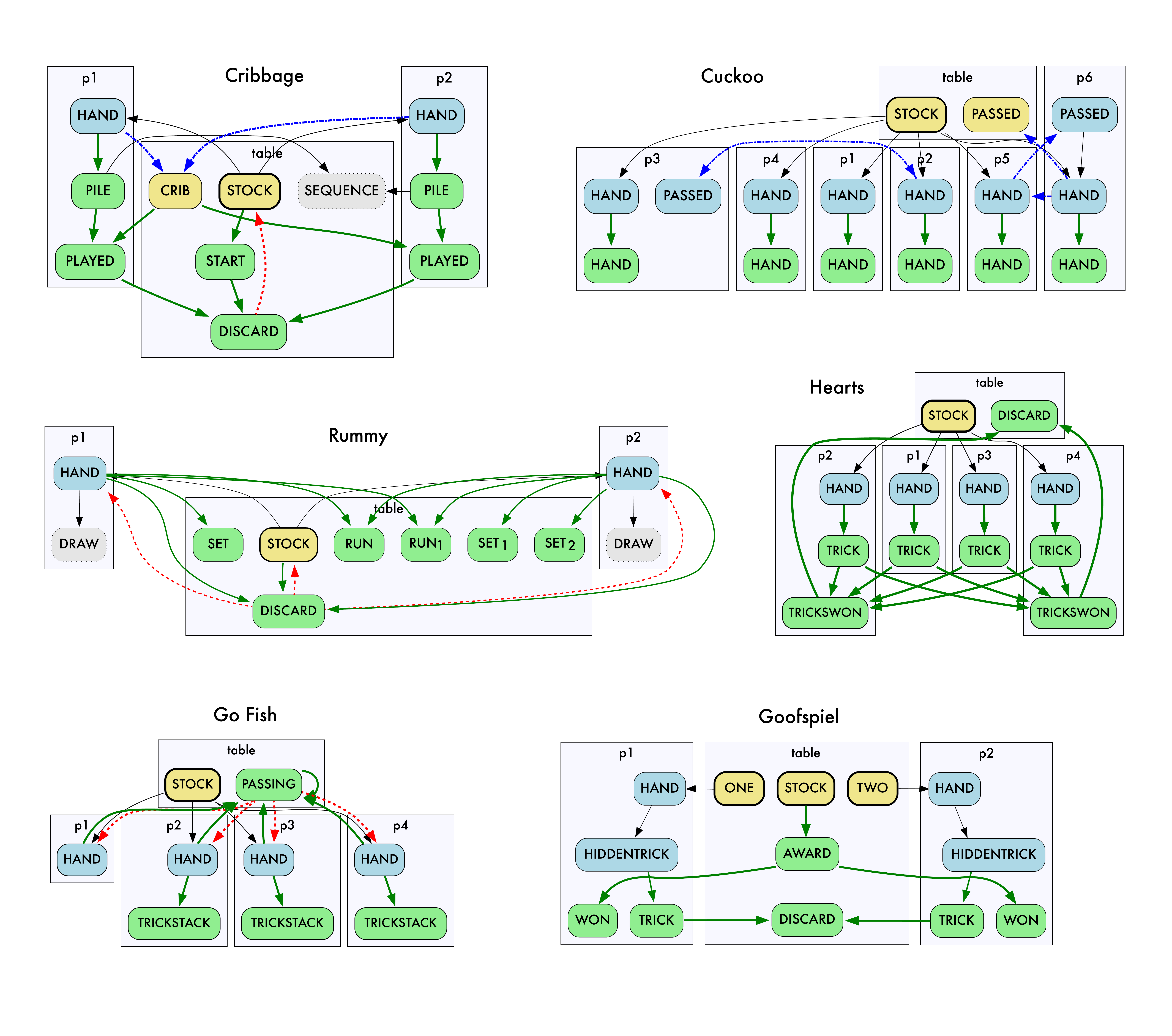}
\caption{Information flow diagrams for six games in Valet.} \label{imperfect}
\end{figure}

Fig.~\ref{imperfect} visualizes information flow in six representative \textit{Valet} games using single random rollouts in CardStock. \textbf{Public} locations are shown in green, \textbf{hidden} locations in yellow, \textbf{private} locations in blue, and memory locations in light gray. Locations containing cards with distinct \textbf{backs} are indicated with bold borders. Rectangles denote ownership (players or table), and directed edges show card movement between locations and resulting visibility changes. \textbf{Taken} information is shown with red dotted edges (public to private/hidden), and \textbf{shared} information with blue dashed edges (private cards transferred to another player).


These diagrams highlight how different games in Valet realize imperfect information. \textsc{Goofspiel} is the only game with multiple card backs, \textsc{Go Fish} and \textsc{Rummy} include public knowledge about taken cards, and \textsc{Cribbage} introduces shared private information through the crib, and in \textsc{Cuckoo}, players are exchanging individual cards. Deduction-based information, common in trick-taking games such as \textsc{Hearts}, is not explicitly visualized. While all games in \textit{Valet} include public and hidden information, the diversity of \textit{Valet} can be seen in the Information column of Table \ref{tab1}, which records other types of information present in each game.

\subsection{Branching Factor}

\begin{figure}[thp]
\includegraphics[width=\textwidth]{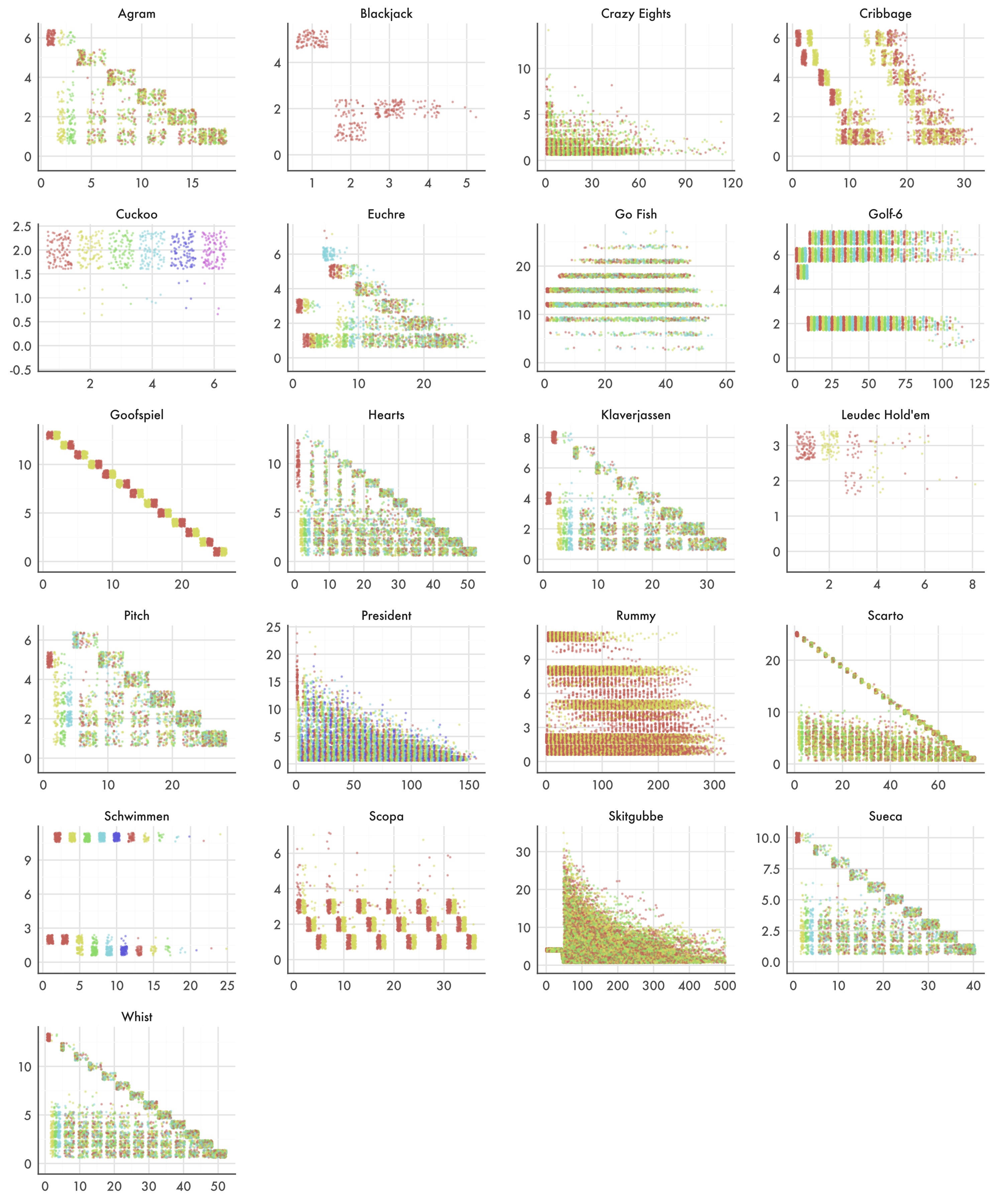}
\caption{The branching factor at each decision point of the game, summarized from 100 random rollouts. Each player is denoted with a different color. Since both axis are integers, an x-y jitter filter is used to help visualize the distributions.} \label{branching}
\end{figure}

The branching factor is a metric used to evaluate the complexity of a game. It quantifies the number of possible actions available to a player in a given situation. A higher branching factor indicates greater game complexity, meaning players must consider more options when making decisions.

Fig.~\ref{branching} illustrates the branching factor (y-axis) for each decision point (x-axis) of the game, summarized over 100 random games. We define a decision point as a specific moment at which a player must choose from a set of possible actions. Following this definition, a player can be part of multiple distinct decision points during one of their turns, resulting in successive actions. 

Most games exhibit a moderate number of actions available at each step, indicating significant yet manageable complexity. The main exceptions are \textsc{Go Fish} and \textsc{President}, which have more complex decision-making processes. However, this is linked to their core mechanics and cannot easily be alleviated. \textsc{Skitgubbe} and \textsc{Scarto} also have high branching factors. In \textsc{Scarto}, this is limited to the first player, who must choose from a large number of cards in their hand. Meanwhile, in the first phase of \textsc{Skitgubbe}, players aim to minimize their hand size for the second phase of the game. However, random agents do not adopt this behavior, resulting in a larger branching factor than those seen in real games.

Interestingly, trick-taking games exhibit a similar branching factor profile, with a strong descending line for the first player, and more limited options for subsequent players due to suit-following constraints. Although trick-taking games are well represented, we observe that each of the other games has a distinct profile, demonstrating diversity within the testbed.




\begin{figure}[th]
\includegraphics[width=\textwidth]{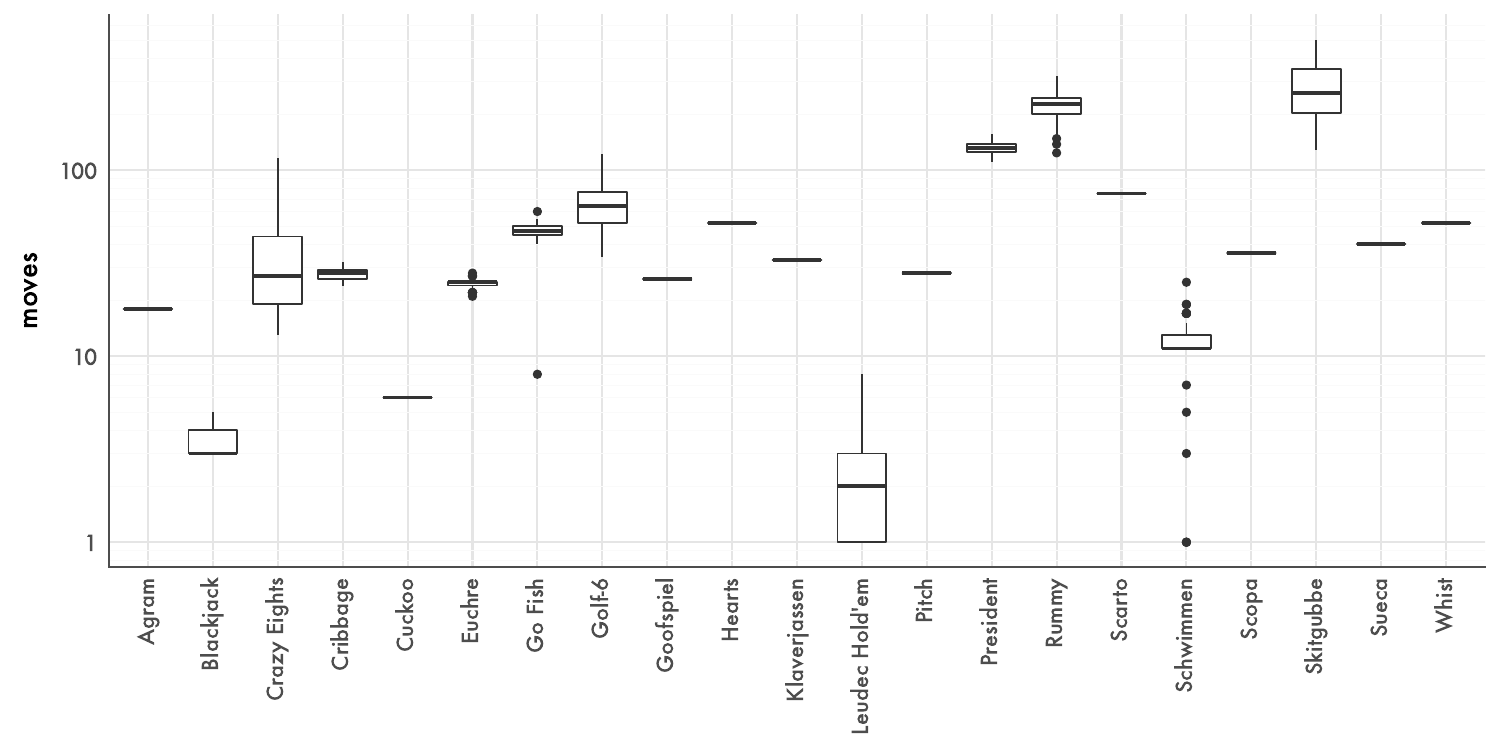}
\caption{The distribution of game lengths over 100 random rollouts for each game. The lengths of the game are on a logarithmic scale.} \label{length} \vspace{-6mm}
\end{figure}

\subsection{Game Length}

Game length is another indicator of complexity: longer games contain more decision points and require planning over longer horizons. Fig.~\ref{length} reports length in decision points. While some games have fixed length (notably most trick-taking games), others are highly variable, with most falling between 10 and 100 decision points.

\textsc{Rummy} and \textsc{Skitgubbe} are outliers with substantially longer games. This is largely due to the availability of low-impact actions that do not meaningfully progress the state, which random agents select more often than human players.


\subsection{Score Distribution}

\begin{figure}[t]
\includegraphics[width=\textwidth]{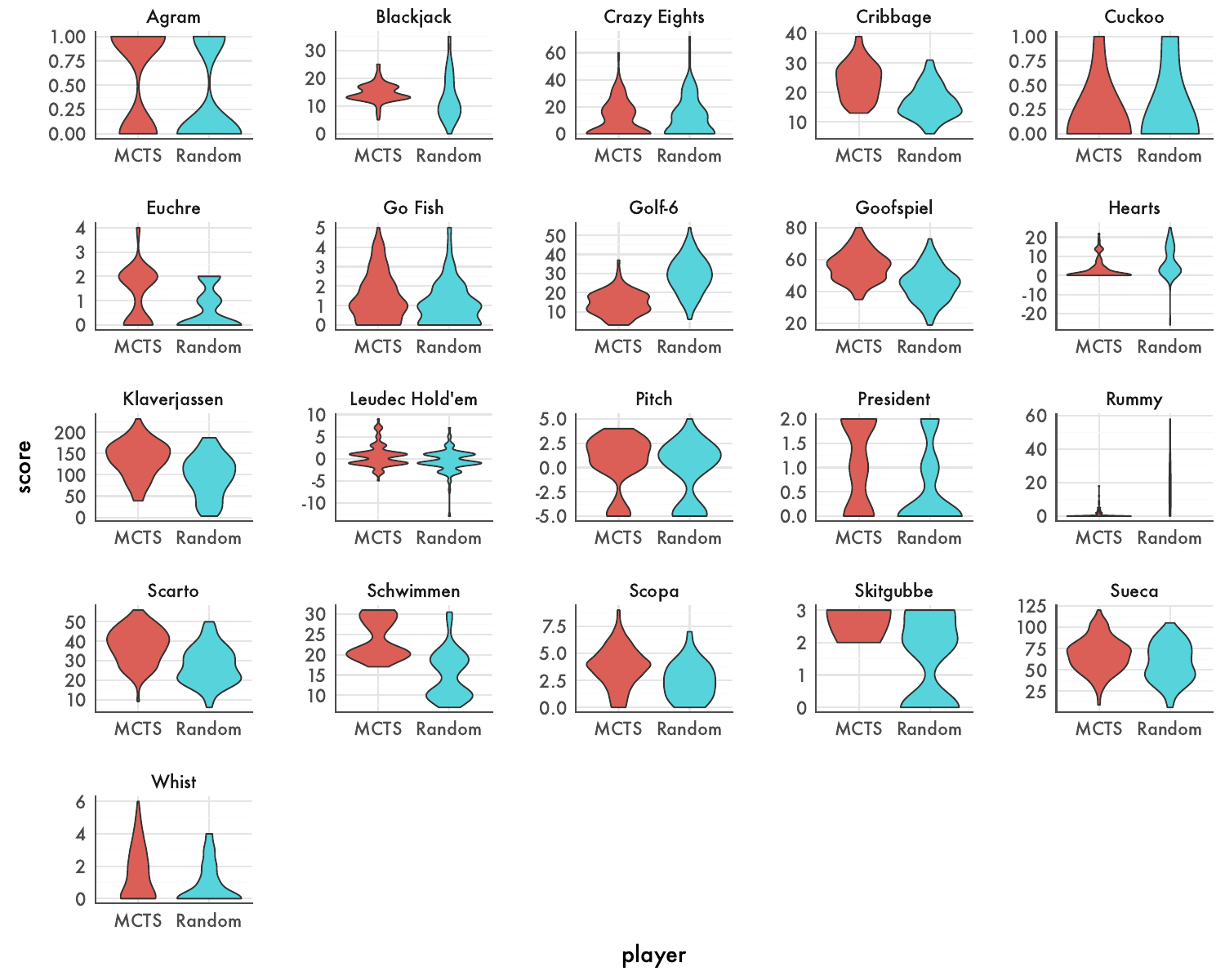}
\caption{A violin plot of the distribution of scores for the first player in each game, comparing MCTS with random. } \label{winrate} \vspace{-6mm}
\end{figure}

Fig.~\ref{winrate} illustrates the diversity of score distributions across \textit{Valet}. We note that all the games have different score distributions. The widest spread occurs in \textsc{Klaverjassen}, where scores range from 0 to over 200 points, while \textsc{Agram} and \textsc{Cuckoo} only produce scores of 0 or 1 due to their one-winner and one-loser win conditions. Some games reward higher scores whereas others are low-score objectives; the MCTS player achieves better outcomes in all games except \textsc{Cuckoo}, possibly because the first player has little opportunity to exploit shared information. \textsc{Rummy} shows the largest performance gap, with MCTS frequently achieving a score of zero, consistent with our earlier observation that effective play requires avoiding inconsequential moves. In contrast, games such as \textsc{Go Fish} and \textsc{Crazy Eights} show little separation between MCTS and random play, likely because CardStock’s determinizations account only for visibility and not for information inferred from action history.


\section{Discussion and Conclusion}

Beyond validating the internal diversity of the testbed, these results highlight how \textit{Valet} can strengthen experimental practice in imperfect-information game AI. For algorithm development, the testbed supports evaluation across varied combinations of mechanics, information structures, branching factors, game lengths, and reward models. This helps distinguish algorithmic components that generalize across genres from those that are tightly coupled to specific mechanics or information patterns. For comparative evaluations, \textit{Valet} mitigates the risk of overfitting conclusions to a small set of canonical games, encouraging performance claims that are robust across a broader spectrum.

For general game-playing systems, the standardized rule sets enable more systematic comparisons between frameworks by controlling for rule interpretation and game variants. This reduces a major source of experimental inconsistency and makes results easier to reproduce and validate, improving cross-paper comparability. Finally, the combination of curated diversity and empirical metrics supports the construction of taxonomies of imperfect-information card games, enabling games to be clustered by mechanics, information structure, or measured complexity, and facilitating analysis of how different AI approaches perform across classes of games.

\textit{Valet} provides a shared testbed of traditional imperfect-information card games for reproducible evaluation of agents and game systems. By enabling experiments across a broad set of games, it supports more robust benchmarking and deeper study of how game properties influence algorithm performance.

A natural extension is a complementary testbed for modern card games, capturing mechanics such as drafting, deck building, asymmetric powers, and cooperative play.

\begin{credits}
\subsubsection{\ackname} This article is based upon work from COST Action CA22145 - GameTable, supported by COST (European Cooperation in Science and Technology).

\subsubsection{\discintname}
The authors have no competing interests to declare that are
relevant to the content of this article.
\end{credits}
%
%
%
\bibliographystyle{splncs04}
\bibliography{mybibliography}

\begin{subappendices}
\renewcommand{\thesection}{\Alph{section}}%

\end{subappendices}

\end{document}